\documentclass[sigconf,screen=true,anonymous=false,bookmarks=false,9pt]{acmart}
\usepackage{varwidth}

\renewcommand\footnotetextcopyrightpermission[1]{} 
\usepackage{lipsum}
\usepackage{titlesec}

\iftrue
\fi

\usepackage{graphicx}
\usepackage{amsmath}
\usepackage{mathtools}
\usepackage{comment}
\usepackage[subrefformat=parens,labelformat=parens]{subfig}
\usepackage{bm}
\usepackage{multirow}
\usepackage{threeparttable,booktabs}
\usepackage{blkarray}
\usepackage{tikz}
\usepackage{balance}
\usepackage{courier}                                       
\usepackage{cleveref}                                      
\usepackage[mathcal]{eucal}
\usepackage[]{algpseudocode}
\algrenewcommand\textproc{\texttt}
\makeatletter\let\float@addtolists\relax\makeatother
\usepackage{algorithm}

\usepackage{filecontents}                                  
\usepackage{pgfplots}
\usepackage{pgfplotstable}
\pgfplotsset{compat=newest}
\usepackage[figuresright]{rotating}

\renewcommand{\vec}[1]{\boldsymbol{#1}}

\theoremstyle{plain}

\theoremstyle{definition}
\newtheorem{mydefinition}{\textbf{Definition}}
\newtheorem{myproblem}{\textbf{Problem}}

\graphicspath{{./figs/}{./}}

\definecolor{NVgreen}{RGB}{118,185,0}
\definecolor{NVblack}{RGB}{0,0,0}
\definecolor{NVlgrey}{RGB}{205,205,205}
\definecolor{NVmgrey}{RGB}{140,140,140}
\definecolor{NVdgrey}{RGB}{94,94,94}

\definecolor{NVemerald}{RGB}{0,133,100}
\definecolor{NVamethyst}{RGB}{93,22,130}
\definecolor{NVintel}{RGB}{0,113,197}
\definecolor{NVgarnet}{RGB}{137,12,88}
\definecolor{NVfluorite}{RGB}{250,194,0}

\begin{document}

\title{
   Large Scale Mask Optimization Via Convolutional Fourier Neural Operator and Litho-Guided Self Training
}

\iftrue
\author{Haoyu Yang}             
\affiliation{ \institution{NVIDIA Corp.} }
\author{Zongyi Li}           
\affiliation{ \institution{NVIDIA Corp. and Caltech} }
\author{Kumara Sastry}               
\affiliation{ \institution{NVIDIA Corp.} }
\author{Saumyadip Mukhopadhyay}             
\affiliation{ \institution{NVIDIA Corp.} }
\author{Anima Anandkumar}           
\affiliation{ \institution{NVIDIA Corp. and Caltech} }
\author{Brucek Khailany}         
\affiliation{ \institution{NVIDIA Corp.} }
\author{Vivek Singh}         
\affiliation{ \institution{NVIDIA Corp.} }
\author{Haoxing Ren}                 
\affiliation{ \institution{NVIDIA Corp.} }
\fi

\begin{abstract}

Machine learning techniques have been extensively studied for mask optimization problems, aiming at better mask printability, shorter turnaround time, better mask manufacturability, and so on.
However, most of these researches are focusing on the initial solution generation of small design regions. 
To further realize the potential of machine learning techniques on mask optimization tasks,
we present a Convolutional Fourier Neural Operator (CFNO) that can efficiently learn layout tile dependencies
and hence promise stitch-less large-scale mask optimization with limited intervention of legacy tools.
We discover the possibility of litho-guided self training (LGST) through a trained machine learning model when solving non-convex optimization problems, 
which allows iterative model and dataset update and brings significant model performance improvement.
Experimental results show that, for the first time, our machine learning-based framework outperforms state-of-the-art academic numerical mask optimizers with an order of magnitude speedup.

\end{abstract}


\maketitle
\renewcommand{\shortauthors}{Yang et al.}

\section{Introduction}
\label{sec:intro}

\begin{figure}
	\centering
	\includegraphics[width=.45\textwidth]{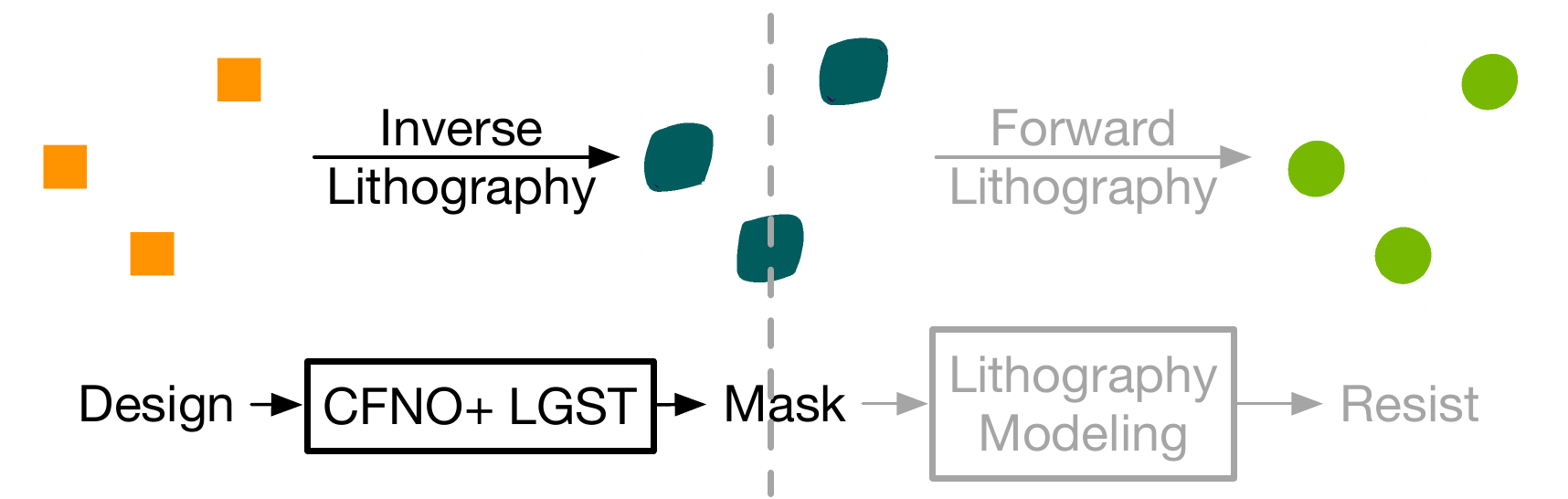}
	\caption{Forward lithography evaluates the resist image on the silicon wafer after the mask going through lithography process.
	Inverse lithography on the other hand is a mask optimization flow that finds the mask such that the resist image after lithography process is as close as the target design.
This paper focuses on the inverse problem.}
	\label{fig:intro}
\end{figure}

Mask optimization is an important step in chip manufacturing flows. 
It tries to find a mask design such that the final pattern on the wafer after lithography process is as close as possible to the target design, as in \Cref{fig:intro}.
Legacy model-based solutions or inverse lithography techniques (ILT) perform mask update through numerical or heuristic optimization by interactively querying lithography models \cite{OPC-DATE2015-Kuang,OPC-DAC2014-Gao,OPC-TCAD2016-Su}.
These solutions are however challenged by the requirements of fast turnaround time.

Recently, machine learning techniques are used to design mask optimization solutions,
such as initial mask generation \cite{OPC-TCAD2020-Yang,OPC-ICCAD2020-DAMO,OPC-ICCAD2020-NeuralILT,OPC-ICCAD2021-Chen}, fast lithography error prediction \cite{OPC-ASPDAC2019-Jiang}, sub-resolution assist feature (SRAF) generation \cite{OPC-TCAD2020-Geng,OPC-ISPD2016-Xu,OPC-DAC2019-Alawieh}, and so on.
GAN-OPC \cite{OPC-TCAD2020-Yang} is the first work using a generative deep learning model to generate initial mask for ILT engines.
DAMO \cite{OPC-ICCAD2020-DAMO} builds an accurate deep learning-based lithography simulator that can guide via/contact layout mask optimization, which reduces mask optimization runtime by a significant amount.
Neural-ILT \cite{OPC-ICCAD2020-NeuralILT} replaces the ILT-guided pretraining technique in GAN-OPC with true numerical lithography engine that further improves mask design quality.
A2-ILT \cite{OPC-DAC2022-Wang} is one of the state-of-the-art academic ILT solution with the aid of reinforcement learning (RL),
where a RL engine is developed to generate optimization constraints that lead to better mask quality.
Similar ideas are also deployed on SRAF generation tasks. 
GAN-SRAF \cite{OPC-DAC2019-Alawieh} is the first work that introduces conditional generative adversarial networks on SRAF generation.
It takes the input of a contact layer and place initial SRAFs in the design, which will be further optimized with commercial tools.

In this paper, we focus on the problem of machine learning-based full-chip mask generation.
Although recent works try to combine machine learning models and legacy solutions to improve the efficiency of legacy mask optimization flows, they have very limited application scenarios due to the following drawbacks:
(1) These machine learning models are relying heavily on legacy OPC engines as in \cite{OPC-TCAD2020-Yang,OPC-ICCAD2021-Chen,OPC-DAC2019-Alawieh}) and ignoring the fact that the training set would be somehow sub-optimal;
(2) These machine learning models are focusing on fix-sized small tile mask optimization and are not considering challenges in large scale design optimization problems.
(3) These machine learning models are back-boned with convolution neural networks that are limited on capturing necessary global information for mask optimization tasks \cite{DL-CVPR2020-Wang}.

\textbf{Fourier Neural Operator As Lithography Learner}. 
Fourier neural operator (FNO) is proposed in \cite{PDE-ICLR2021-Li} as a partial differential equation (PDE) solver.
FNO takes an embedding tensor as the input and performs global information mixing in the Fourier domain followed by non-linearity in the input domain.
FNO also resembles the approximated lithography modeling that offers the opportunity to learn lithography behavior efficiently \cite{DFM-B2011-Ma,DFM-DAC2022-Yang}.
An example is DOINN \cite{DFM-DAC2022-Yang} that employs an optimized FNO layer for fast and accurate lithography modeling.
DOINN is designed to support any-sized input without loss of accuracy on lithography modeling. 
Mask optimization, on the other hand, is a more challenging task which requires effective global information acquisition because the optimization results have long range dependency.
This is because mask optimization is an inverse flow of forward lithography, where the resist image of a location is determined by the contexts of its surrounding region. 
Thus, a shape modification in the mask image will affect the optimization of its neighbours and therefore makes the mask optimization problem a global optimization flow.
Failing to capture these design long range dependencies will certainly result in additional cost to fix the mask when performing tile-based full-chip mask optimization \cite{OPC-SPIE2019-Pang}. 
Therefore, mask optimization problem by nature requires high resolution and large tile inputs to preserve enough information. 
This however poses extreme computing cost in standard FNO due to multiple Fourier Transforms in the data pipeline.

\textbf{Convolutional Fourier Neural Operator.}
To address the challenges of machine learning-based mask optimization problems, 
we propose a customized model termed as convolutional FNO (CFNO). 
CFNO introduces token shared FNO unit to avoid Fourier Transform on large inputs. 
Equipped with token-wise convolution layer, CFNO achieves effective local-global mixing for the purpose of large scale mask optimization tasks.

\textbf{Litho-Guided Self Training.}
In most of the machine learning applications (e.g.~classification, segmentation, object detection and so on), a model is trained with data-label pairs, 
where the labels are correct or optimal. 
Mask optimization tries to find a solution, such that some manufacturing-aware objectives are minimized.
However, the non-convexity of mask optimization problems makes it impossible to obtain optimal labels for model training.
\textit{Thus, the machine learning generated results are not necessarily bad even they differ from the training golden value by a significant level.} 
The fact gives us an opportunity to update the training set and the machine learning model alternatively for better performance. 
We denote the procedure as litho-guided self training (LGST).
Unlike traditional self training algorithms \cite{reichart2007self,OPC-ISPD2021-Jiang}, where the machine learning model is applied to unlabeled data to create labeled instances for further training, 
LGST tries to improve the label quality of existing labeled dataset and pursues a higher training set quality.
We will show in the experiments how this fact benefits our framework with significantly improved mask optimization quality.

\textbf{Major Contributions.} Our major contributions are summarized as follows:
\begin{itemize}
    \item We develop a CFNO structure as the efficient mask optimization engine through token-shared FNO unit and token-wise convolution operation.
    \item We present the idea of litho-guided self training when developing machine learning-based solution for mask optimization tasks.
    \item We conduct extensive experiments on CFNO-backbone and LGST. \textbf{For the first time, a pure machine learning framework achieves even better results ($3\times$ and $100\times$ smaller EPE violation) than numerical optimization-based solution with $600\times$ speedup.}
\end{itemize}

\section{Preliminaries}
\label{sec:prelim}
This section introduces basic terminologies related to mask optimization and machine learning. 
Throughout the paper, we use lowercase letters (e.g.~$x$) for scalars, bold lowercase letters for vectors (e.g.~$\vec{x}$), bold uppercase letters (e.g.~$\vec{X}$) for matrices or tensors.

Forward lithography modeling is developed to estimate the lithograph behavior in real manufacturing flows. 
Singular value decomposition approximation \cite{DFM-B2011-Ma} is the most commonly used approach for lithography modeling, which can be expressed as:
\begin{align}
	\vec{I}(m,n)=\sum_{k=1}^{N^2} \alpha_k |\vec{h}_k(m,n) \otimes \vec{M}(m,n) |^2,
	\label{eq:svd-all}
\end{align}
where $\otimes$ denotes the convolution operation, $\vec{M}(m,n)$ indicates the mask, $\vec{I}(m,n)$ is the corresponding aerial image, $\vec{h}_k(m,n)$'s are lithography kernels and $\alpha_k$'s are kernel related coefficients.
\Cref{eq:svd-all} will be followed with constant threshold resist modeling:
\begin{align}
    \vec{Z}(m,n)=
    \begin{cases}
	0, & \text{if}\ \vec{I}(m,n)<D_{th}, \\
	1, & \text{otherwise},
	\end{cases}
\end{align}
where $D_{th}$ is some predefined resist threshold value and $\vec{Z}(m,n)$ represents the resist image.

\begin{figure}[tb!]
	\centering
	\includegraphics[width=.38\textwidth]{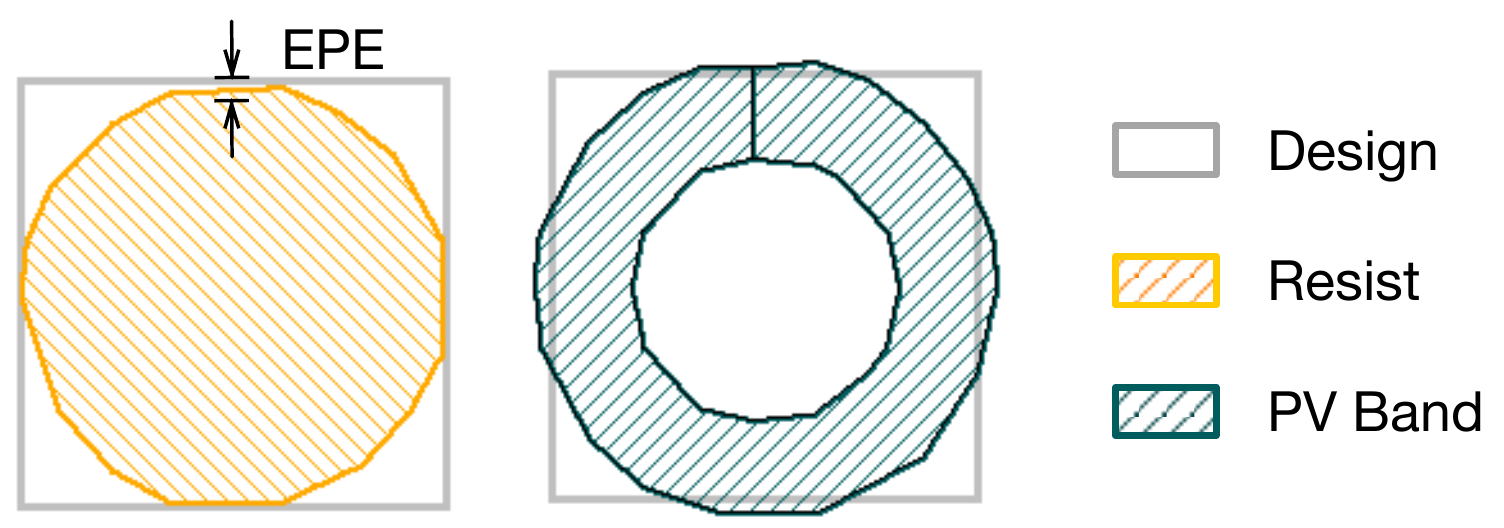}
	\caption{Mask quality measurements.}
	\label{fig:measure}
\end{figure}

Mask optimization (MO) is a problem to find a proper mask $\vec{M}$ associated with a design $\vec{Z}_t$, such that the difference between the resist image $\vec{Z}$ after the forward lithography modeling and the design is minimized.
In literature, there are many evaluation metrics used to estimate the quality of mask optimization solutions. 
Well accepted ones are edge-displacement-error (EPE) violations, mean square error (MSE) and process variation band (PVB) area.
EPE and PVB are depicted in \Cref{fig:measure}.

\begin{mydefinition}[EPE Violation\cite{OPC-ICCAD2013-Banerjee}]
EPE is measured as the geometric distance between the target edge and the lithographic contour printed at the nominal condition. 
If the EPE measured at a point is greater than certain tolerance value, we call it an EPE violation.

\end{mydefinition}

\begin{mydefinition}[MSE]
\label{def:mse}
MSE measures the pixel-wise difference between the design and the resist image as in:
\begin{align}
	\text{MSE}=||\vec{Z}-\vec{Z}_t||_F^2.
\end{align}
\end{mydefinition}

\begin{mydefinition}[PVB Area\cite{OPC-ICCAD2013-Banerjee}]
This is evaluated by running lithography simulation at different corners on the final mask solution.
Once run, a process variation band metric will be defined as the XOR of all the contours.
The total area of the process variation band is defined as PVB Area.
\end{mydefinition}

These evaluation metrics are equally important.
We can see that both EPE and MSE directly measures the error between resist images and designs.
The only difference is that MSE evaluates the resist image in a more general perspective while EPE focuses on critical measurement points.
On the other hand, PVB is related to the robustness of masks subject to potential process variations.
With these evaluation metrics, we can accordingly formulate the machine learning-based mask optimization (MLMO) problem as follows:

\begin{myproblem}[MLMO]
	Given a set of designs $\mathcal{Z}_{tr}=\{\vec{Z}_{tr,1}^\ast, \vec{Z}_{tr,2}^\ast, ..., \vec{Z}_{tr,n}^\ast\}$ and their corresponding masks from some mask optimization engine $\mathcal{M}_{tr}=\{\vec{M}_{tr,1}, \vec{M}_{tr,2}, ..., \vec{M}_{tr,n}\}$,
	our objective is to build a machine learning model $f(\cdot, \vec{W})$ such that for new designs $\mathcal{Z}_{te}^\ast=\{\vec{Z}_{te,1}^\ast, \vec{Z}_{te,2}^\ast, ..., \vec{Z}_{te,m}^\ast\}$,
	$f$ will produce the corresponding masks $\mathcal{M}_{te}=f(\mathcal{Z}_{te}^\ast) =\{\vec{M}_{te,1}, \vec{M}_{te,2}, ..., \vec{M}_{te,m}\}$ and the mask quality measured in terms of EPE Violation, MSE and PVB Area is optimized.
\end{myproblem}

\section{The Framework}
\label{sec:alg}
This section will cover the details of our mask optimization framework that includes CFNO-backboned neural network design and training with LGST.

\subsection{Convolutional Fourier Neural Operator}

\subsubsection{FNO Basis:}
\begin{figure}[tb!]
	\centering
	\includegraphics[width=.48\textwidth]{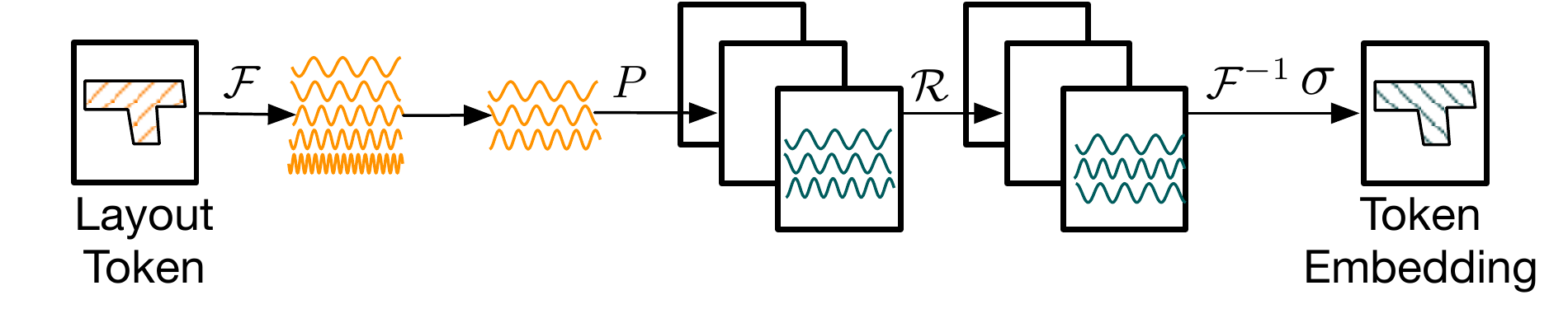}
	\caption{Data pipeline of the Fourier Neural Operator.}
	\label{fig:fno}
\end{figure}

Our framework starts from the Fourier Neural Operator, which defines a kernel $\kappa$ integral at some token $g$:
\begin{align}
    u = \sigma (\mathcal{F}^{-1}(\mathcal{F}\kappa \cdot \mathcal{F}v)),
    \label{eq:fno}
\end{align}
where $\mathcal{F}$ and $\mathcal{F}^{-1}$ denote the Fourier Transform and the Inverse Fourier Transform, $u,v$ represent continuous functions, and $\sigma$ is some activation function.
Let $v$ be the input image token, \Cref{eq:fno} can be rewritten into the discrete form:
\begin{align}
    \vec{U} = \sigma (\mathcal{F}^{-1}(\mathcal{F}(\vec{K}) \cdot \mathcal{F}(\vec{V}))),
    \label{eq:dfno}
\end{align}
where $\vec{V},\vec{K},\vec{U} \in \mathbb{R}^{h\times w}$ represent token image, global convolution kernel and token embedding, respectively. 
It should be noted that \Cref{eq:dfno} is equivalent to
\begin{align}
   \vec{U} = \sigma (\vec{K} \otimes \vec{V}), 
\end{align}
which resembles the computation inside \Cref{eq:svd-all} and hence makes FNO a preferred lithography learner. 
We do not explicitly train the global convolution kernel $\vec{K}$ for the sake of computing overhead. 
Instead, a frequency mixing weight $\vec{W} = \mathcal{F}(\vec{K}) \in \mathbb{C}^{h \times w}$ is directly introduced.
\Cref{eq:dfno} therefore becomes,
\begin{align}
    \vec{U} = \sigma (\mathcal{F}^{-1}(\vec{W} \cdot \mathcal{F}(\vec{V}))).
    \label{eq:dfno-1}
\end{align}
Because the FNO is designed for global information acquisition, in real implementation, only low frequency components are kept for $\mathcal{F}(\vec{V})$.
Also, $\mathcal{F}(\vec{V})$ is mapped to a higher dimension through channel-lifting (see \cite{DFM-DAC2022-Yang}) before convolving with the global convolution kernel.
For simplicity, these settings are not reflected in the equations.
The detailed data pipeline in FNO is depicted in \Cref{fig:fno}.

In the original FNO design \cite{PDE-ICLR2021-Li}, the size of $\vec{V}$ determines the receptive field of the global convolution.
This, however, poses us great challenges when dealing with data with long range spatial dependency.
Mask optimization is a representative example:
(1) \Cref{eq:svd-all} indicates that mask optimization results of a shape is affected by the context information of its neighbours within a reasonable radius,
which requires a minimal dimension of $\vec{V}$.
(2) Mask optimization should be conducted based on large tile unit to reduce efforts when resolving boundary inconsistency (stitching issue) \cite{OPC-SPIE2019-Pang}. 
Both these facts require the computation of Fourier Transforms on very large input and makes FNO less efficient.
To address these concerns, we propose the concept of the \textit{Convolutional Fourier Neural Operator}.

\subsubsection{CFNO Design:}
Vision transformer (ViT) \cite{ViT} is a family of a structure for rich contextual representation learning that considers images as a token sequence.
Image tokens will then be fed into token mixers for subsequent feature embedding.
Inspired by the success of ViT and token-mixing, we develop the Convolutional Fourier Neural Operator for efficient global layout token embedding and resolving layout long range dependency caused stitching issue.

\begin{figure}[tb!]
	\centering
	\includegraphics[width=.44\textwidth]{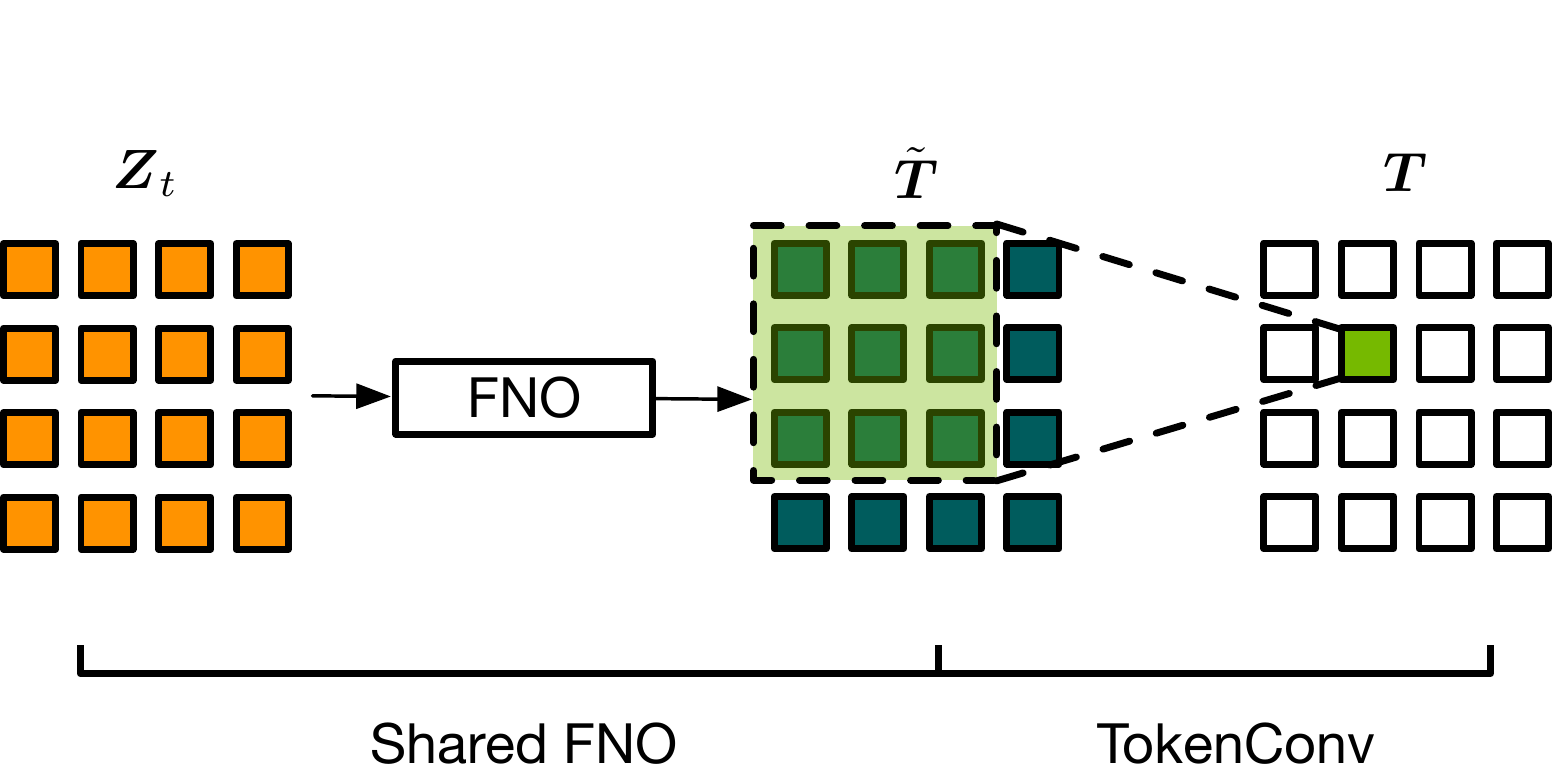}
	\caption{Convolutional Fourier Neural Operator.}
	\label{fig:cfno}
\end{figure}

The core components of CFNO are a token-shared FNO and a token-wise convolution operator as depicted in \Cref{fig:cfno}.
For the token shared FNO, we used the same pipeline as in \Cref{eq:dfno-1} and \Cref{fig:fno}.
The only difference is that the FNO is applied on layout tokens instead of the entire layout image.
Given a design layout image $\vec{Z}_t \in \mathbb{R}^{H \times W}$, we first divide it into non-overlapped patches, referred as tokens:
\begin{align}
    \vec{Z}_t = 
    \begin{bmatrix}
    \vec{Z}_{t,1,1} & \vec{Z}_{t,1,2} & ... & \vec{Z}_{t,1,n} \\
    \vec{Z}_{t,2,1} & \vec{Z}_{t,2,2} & ... & \vec{Z}_{t,2,n} \\
    ... & ... & ... & ... \\
    \vec{Z}_{t,m,1} & \vec{Z}_{t,m,2} & ... & \vec{Z}_{t,m,n} 
    \end{bmatrix},
\end{align}
where $\vec{Z}_{t,i,j} \in \mathbb{R}^{k \times k}$'s are layout tokens, $H=mk$ and $W=nk$.
We define the shared FNO $f(\cdot;\vec{W}_1)$ to get the first level token embedding:
\begin{align}
    \tilde{\vec{T}}_{i,j} = f(\vec{Z}_{t,i,j};\vec{W}_1), i=1,2,...,m, j=1,2,...,n,
    \label{eq:cfno-1}
\end{align}
where $\vec{W} \in \mathbb{C}^{k \times k \times d}$ and $d$ denotes the lifted channel number.
Obviously, \Cref{eq:cfno-1} can be finished efficiently through batch processing and a smaller $k$ indicates a shared FNO with fewer trainable parameters.
However, this token-shared approach scarifies the ability of global information acquisition for model size.

To tackle this concern, we further introduce the second level token embedding via a token-wise convolution parametered with $\vec{W}_2 \in \mathbb{R}^{(2s+1) \times (2s+1)}$:
\begin{align}
    \vec{T}_{i,j} = \sum_{t_x=-s}^{s} \sum_{t_y=-s}^{s} \vec{W}_{2}[i+t_x, j+t_y] \cdot \tilde{\vec{T}}_{i+t_x,j+t_y},
    \label{eq:cfno-2}
\end{align}
which finally formulates the layout global embedding:
\begin{align}
    \vec{T}= 
\begin{bmatrix}
	\vec{T}_{1,1} & \vec{T}_{1,2} & ... & \vec{T}_{1,n} \\
	\vec{T}_{2,1} & \vec{T}_{2,2} & ... & \vec{T}_{2,n} \\
	... & ... & ... & ... \\
	\vec{T}_{m,1} & \vec{T}_{m,2} & ... & \vec{T}_{m,n} 
\end{bmatrix}.
\end{align}
\Cref{eq:cfno-2} defines how tokens at different spatial locations are mixed and hence addresses token boundary inconsistency issue and long-range dependency requirements.

\Cref{tab:fnovscfno} compares CFNO and FNO from the perspective of computing complexity and data flow,
where $N=HW=mnk^2$ is the total size of $\vec{Z}_t$, $d$ is the number of channels lifted in FNO, $k$ is the token size, and $mn$ represents the total number of tokens in the design layout image.
Usually we have $s \ll k$, which grants CFNO both computing and memory efficiency.
It should also be noted that \textit{CFNO enables training and inference on any-sized input without further manipulation}.

\begin{table}[tb!]
\centering
\caption{Comparison between FNO and CFNO.}
\label{tab:fnovscfno}
\begin{tabular}{c|cc}
\toprule
Operator  & FNO & CFNO \\ \midrule
FLOPS     & $N\log N + Nd^2$    & $N \log k^2 +s^2mnd^2$      \\
Parameter & $Nd^2$    & $s^2d^2$     \\ 
DataFlow  &$\mathcal{F}-\text{Linear}-\mathcal{F}^{-1}$ & $\mathcal{F}-\text{Linear}-\mathcal{F}^{-1}-\text{Conv}$\\ \bottomrule
\end{tabular}
\end{table}

\subsubsection{Architecture Summary:}
We can observe that CFNO is defined by two key hyper-parameters: the token size $k$ and the token-wise convolution kernel size $2s+1$.
These two parameters work together to determine how global layout information is acquired.
Inspired by the inception module for multi-scale feature learning \cite{GoogleNet}, we design our final network architecture with four embedding paths.

\begin{figure}[tb!]
	\centering
	\includegraphics[width=.3\textwidth]{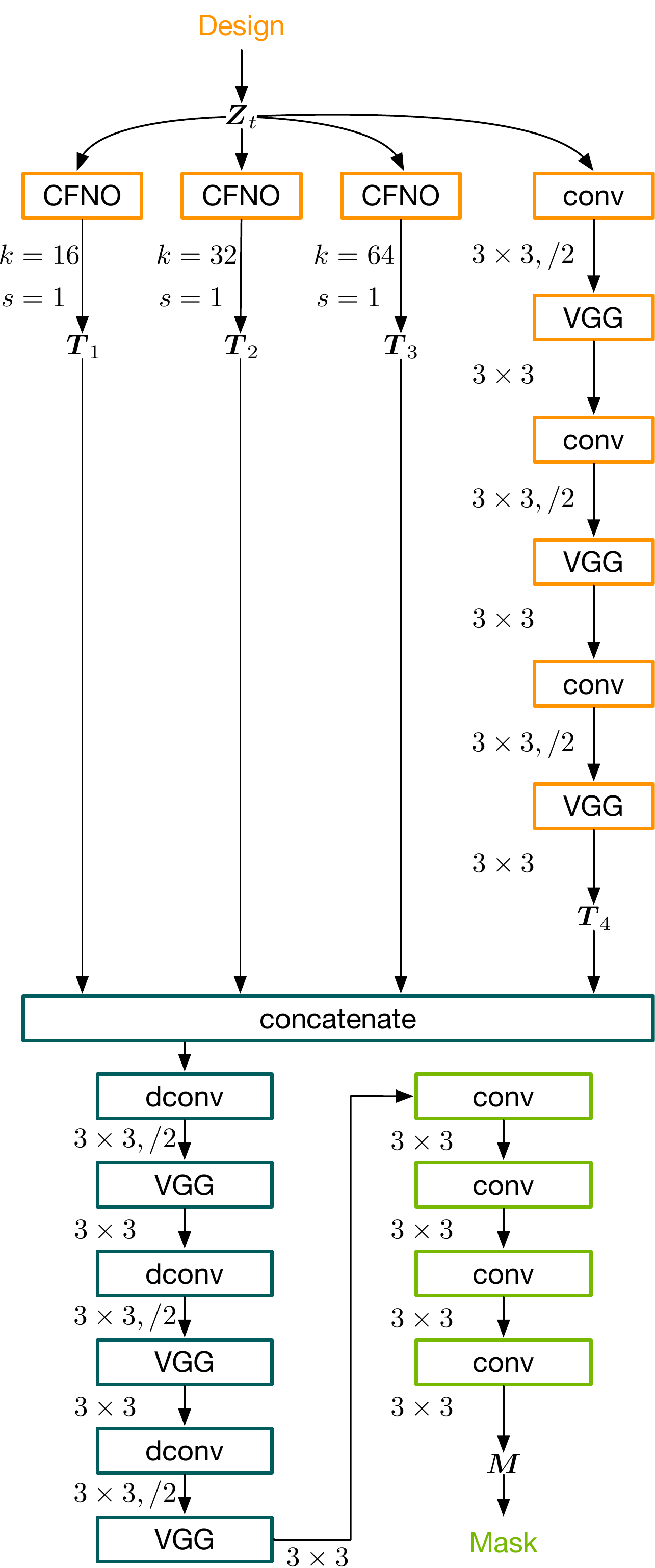}
	\caption{The final structure of the CFNO-based mask optimizer.  \texttt{conv} and \texttt{dconv} represent convolution and transposed convolution layers. 
	\texttt{VGG} denotes a stacked convolution block as proposed in \cite{VGG}. 
	$3\times 3$ indicates the convolution kernel size and $/2$ represents a stride of 2. $k,s$ define the layout token size and the token-wise convolution kernel size, respectively.}
	\label{fig:allnet}
\end{figure}

As shown in \Cref{fig:allnet}, three paths are regular CFNO units with different token size for multi-scale token embedding. 
The last one contains several groups of convolution layers, which can also be viewed as a special case of CFNO with $k=1$.
We have two motivations to design the fourth convolution path: 
(1) Discrete Fourier Transforms assume periodic image inputs for global convolution operation, which is not necessarily true for layout images. 
We therefore include this convolution path for the compensation of boundary information. Similar settings have also been discussed in \cite{UFNO,PDE-ICLR2021-Li}.
(2) In each shared FNO, high frequency coefficients are truncated out to focus on global information acquisition.
These high frequency components are however important is mask learning, because pixel-level changes on masks will result in great change on wafer images.
Recent research has discovered that convolution layers are suitable for high frequency knowledge understanding \cite{DL-CVPR2020-Wang}, and this motivates us the design of a convolution path to compensate high frequency information loss. 

Once we get the token embedding from the four learning paths, we perform one-step aggregation to gather all learned information.
This will be followed by a series of convolution and transposed convolution layers to generate masks. 

 \subsection{Litho-Guided Self Training}
This section focuses on the mask optimization-dedicated training algorithm. 
First we will discuss some key characteristics of the MLMO problem.

\subsubsection{Learning From a Mask Optimizer:}
Most of the discriminative machine learning tasks (classification, segmentation, object detection, and so on) are trying to build some machine learning model to fit a group of observations that will be treated as data-label pairs,
which is also the scenario of most MLMO solutions.
However, the labels (referred as optimized masks) are usually obtained through numerical mask optimizers running OPC or ILT \cite{OPC-DAC2014-Gao,OPC-ICCAD2021-Chen,OPC-DATE2015-Kuang}
which poses the following concerns:
\begin{itemize}
	\item Mask optimizer generated solutions are most likely not optimal because the mask optimization problem itself is non-convex.
	\item It is time consuming to obtain an optimized mask from a given design.
\end{itemize}
Thus, during inference time of a machine learning model trained with these design-mask pairs, 
we cannot determine the quality of a machine learning generated mask by simply measuring its difference from the numerical optimizer solution and we need a lithography checker to evaluate the mask quality.

\subsubsection{Machine Learning Can Do Better:}
Above discussion reveals a gap between MLMO and legacy mask optimization problems. We now want to ask a question: 

\textit{What information can machine learning model learn from the less-optimal labels?}

\begin{figure}[tb!]
\centering 
\subfloat[Design]{\includegraphics[width=.14\textwidth,trim={750 750 0 0},clip]{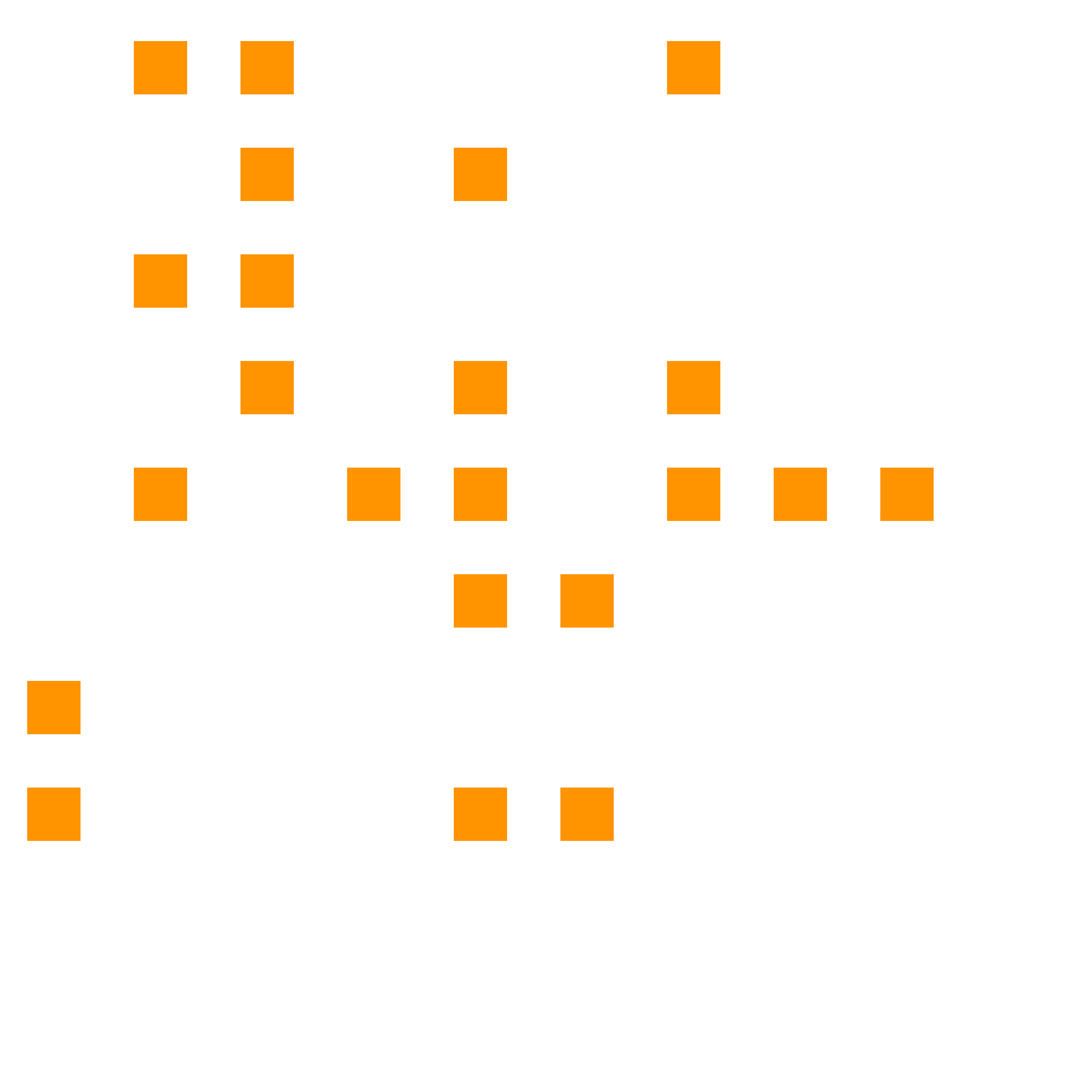} \label{fig:orexample-design}} \hspace{0.15cm}
\subfloat[ILT-Mask]{\includegraphics[width=.14\textwidth,trim={750 750 0 0},clip]{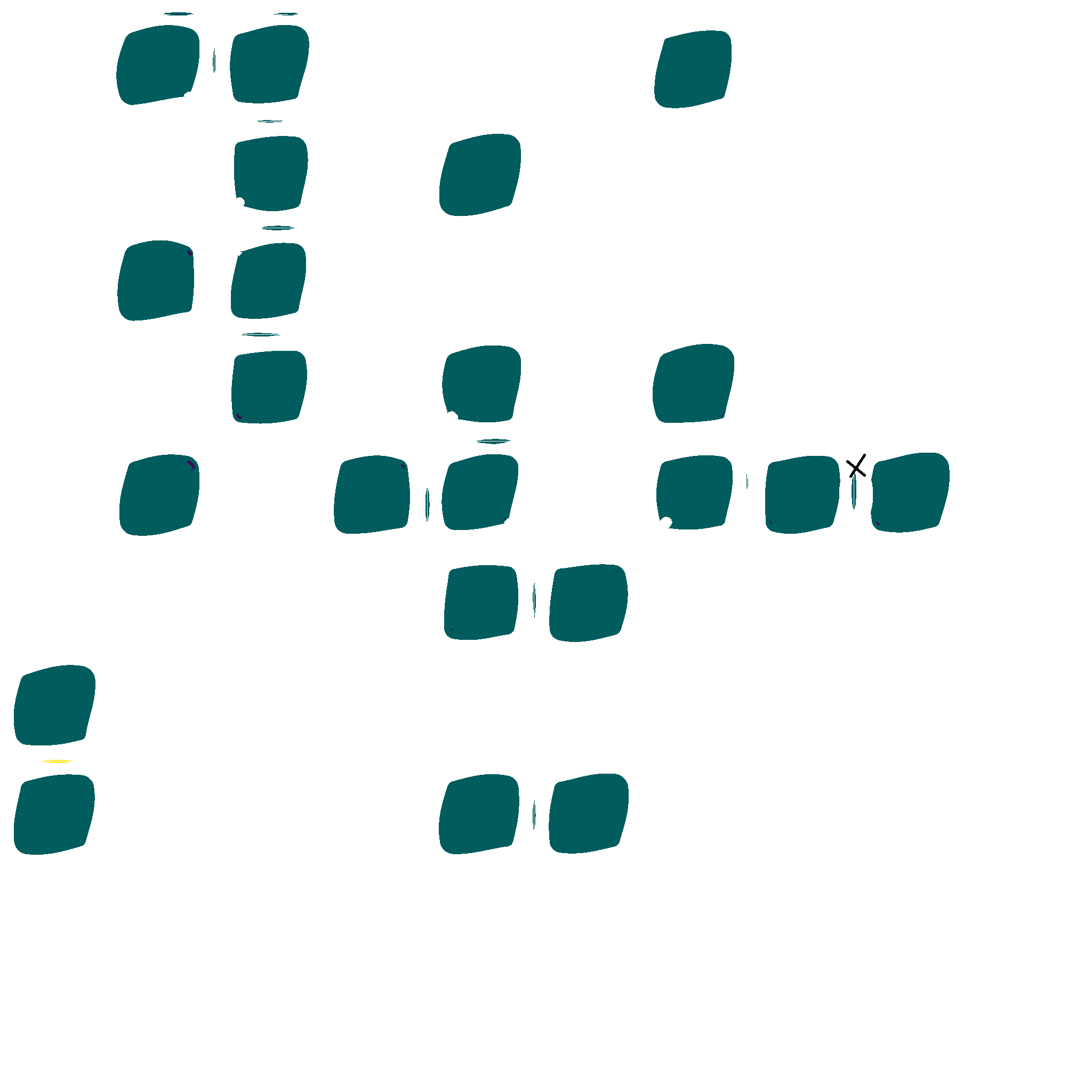} \label{fig:orexample-iltm}} \hspace{0.05cm}
\subfloat[ILT-Resist]{\includegraphics[width=.14\textwidth,trim={750 750 0 0},clip]{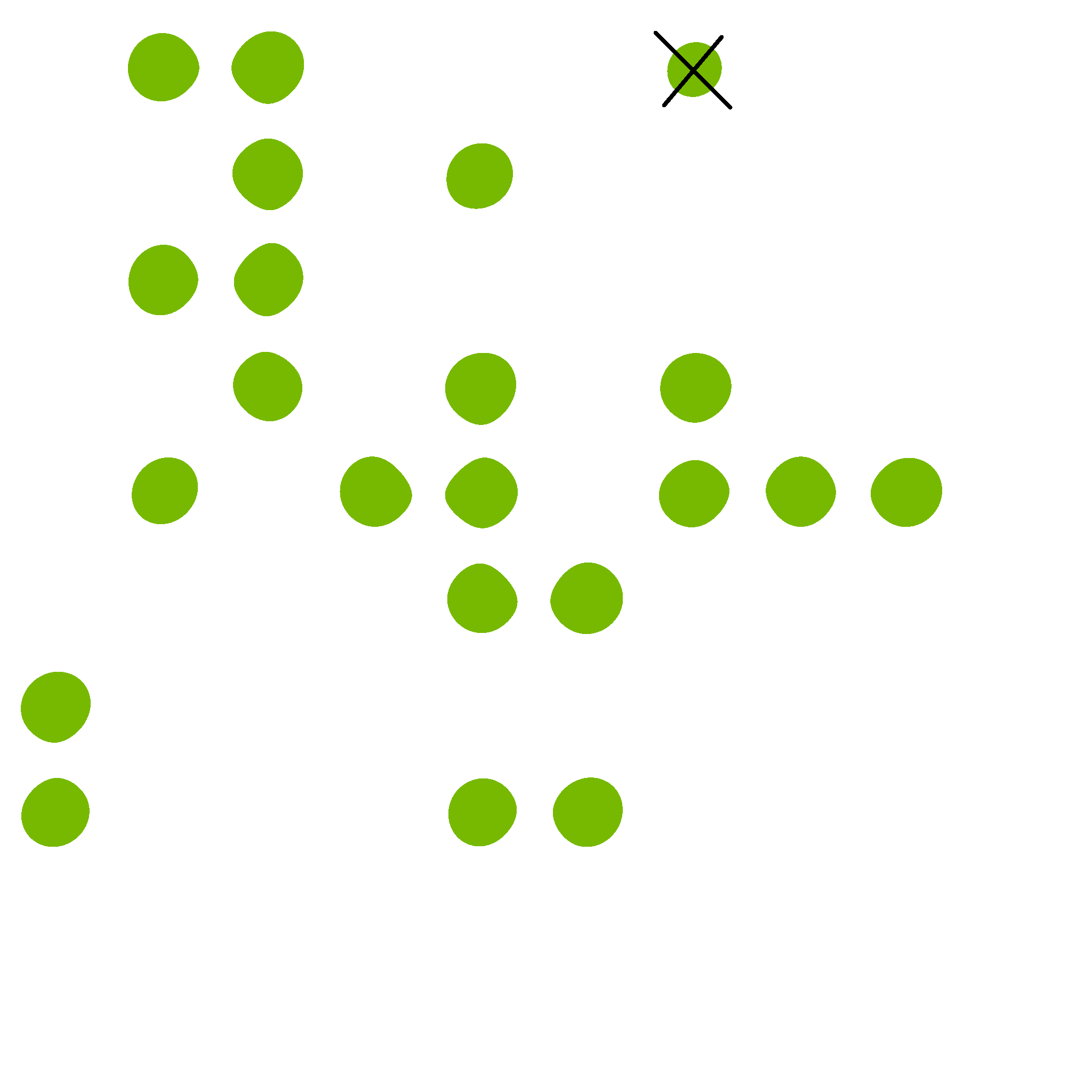} \label{fig:orexample-iltr}}\\
\hspace{2.75cm}
\subfloat[ML-Mask]{\includegraphics[width=.14\textwidth,trim={750 750 0 0},clip]{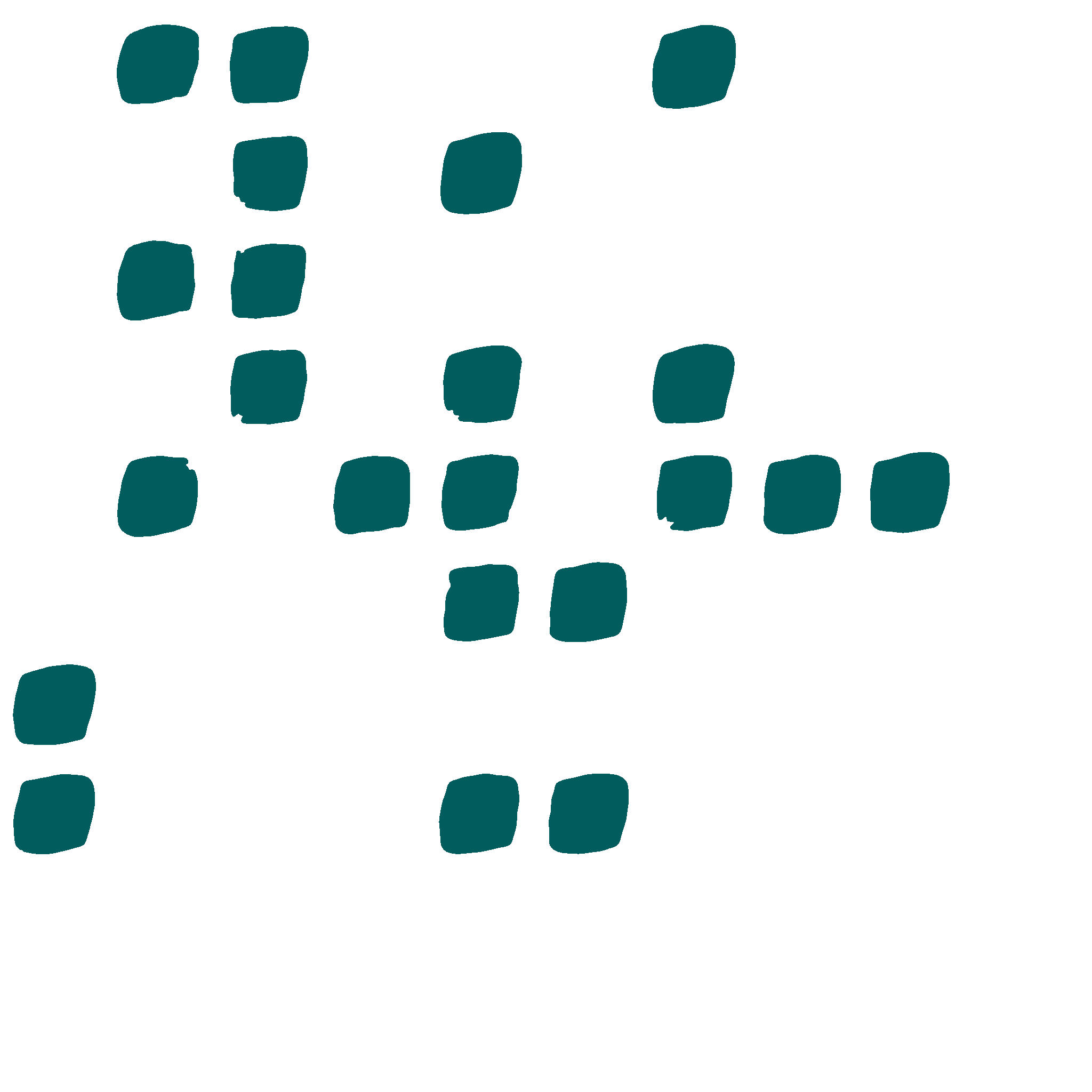} \label{fig:orexample-mlm}}\hspace{0.05cm}
\subfloat[ML-Resist]{\includegraphics[width=.14\textwidth,trim={750 750 0 0},clip]{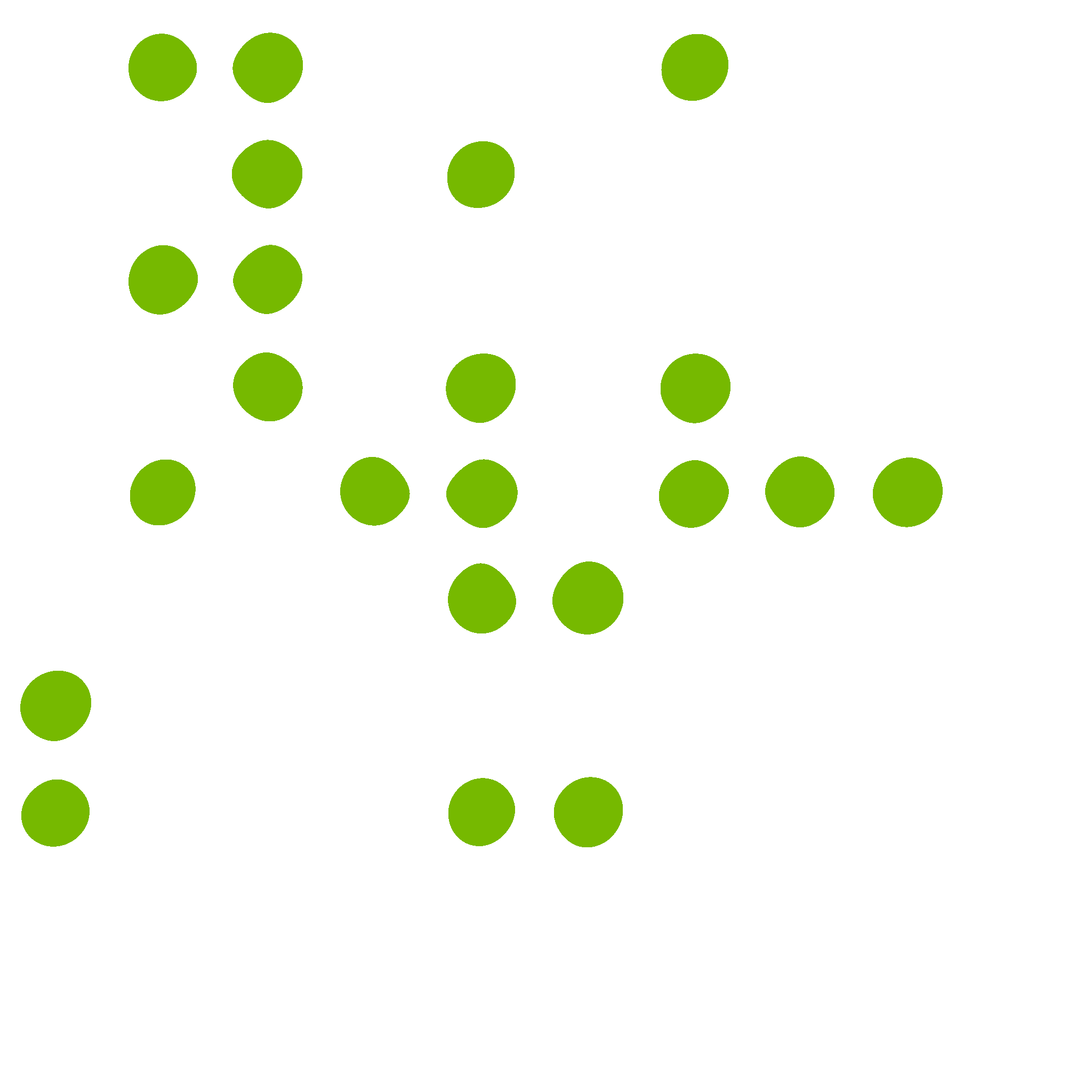} \label{fig:orexample-mlr}}
\caption{Machine learning can do better on mask optimization tasks. 
(a) Part of a design containing via arrays. 
(b) Mask generated by the levelset ILT engine. 
(c) Nominal resist image from the ILT-Mask.
(d) Mask generated by the machine learning model.
(e) Nominal resist image from the ML-Mask.}
\label{fig:orexample}
\end{figure}

We answer this question with \Cref{fig:orexample}.
\Cref{fig:orexample-design} is a design instance from the training set which will be fed into the neural network.
\Cref{fig:orexample-iltm} corresponds to the mask generated through a levelset ILT optimizer.
Compared to the neural network generated mask (\Cref{fig:orexample-mlm}), ILT-Mask contains rule-violation artifacts (crossed in the figure).
We can also observe that isolated resist image is much smaller than the target and the shape in ML-Resist image.
This is because the ILT is a gradient-based solution to minimize the pixel-wise difference between the simulated contours and the design target.
If shapes in a design are unevenly distributed, the low density regions will have smaller gradient and thus cannot be optimized efficiently.

\begin{figure}[tb!]
    \centering
    \subfloat[Design]{\includegraphics[width=.14\textwidth,trim={525 500 250 272},clip]{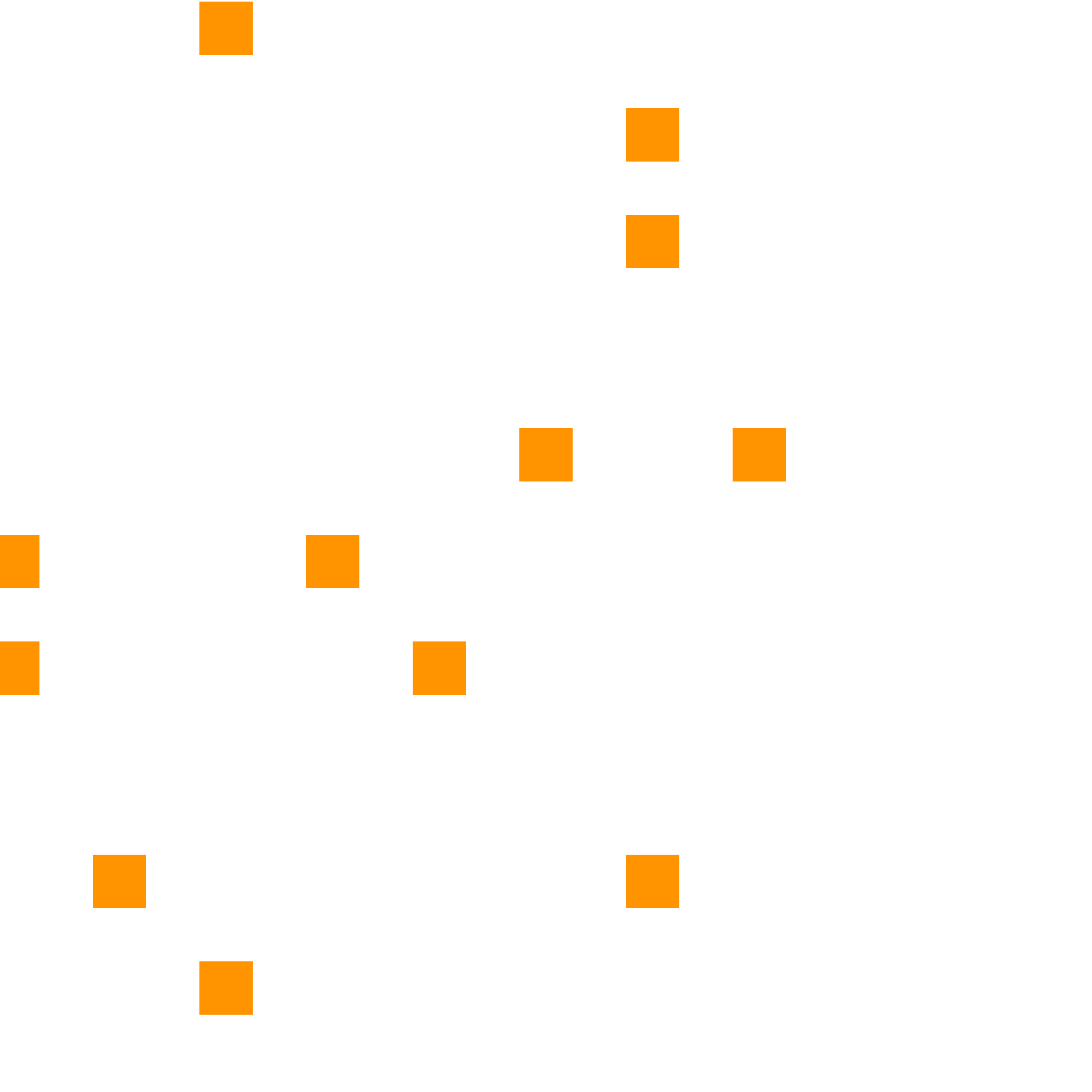} \label{fig:goodexample-design}} \hspace{0.15cm}
    \subfloat[ILT-Mask]{\includegraphics[width=.14\textwidth,trim={525 500 250 272},clip]{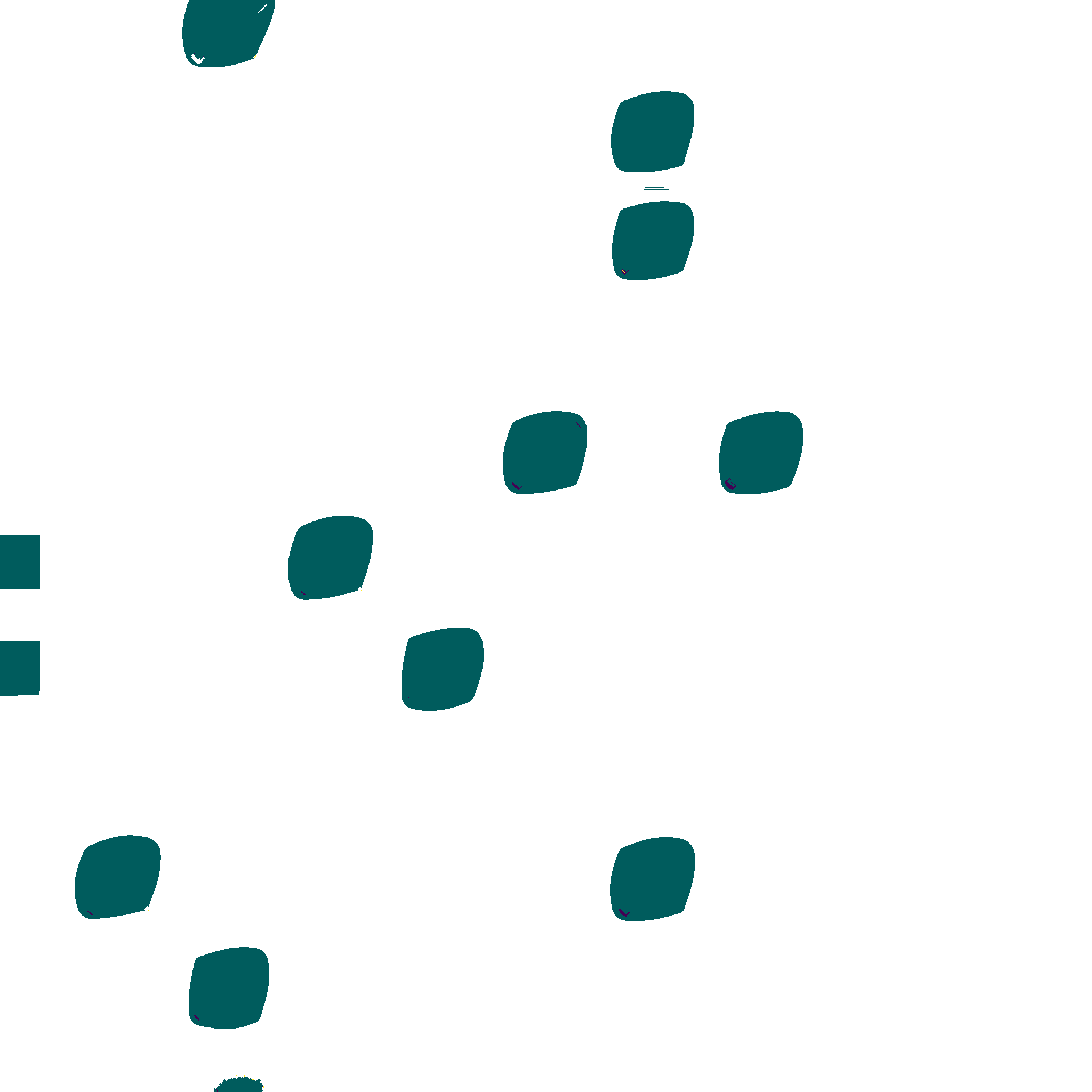} \label{fig:goodexample-iltm}} \hspace{0.15cm}
    \subfloat[ILT-Resist]{\includegraphics[width=.14\textwidth,trim={525 500 250 272},clip]{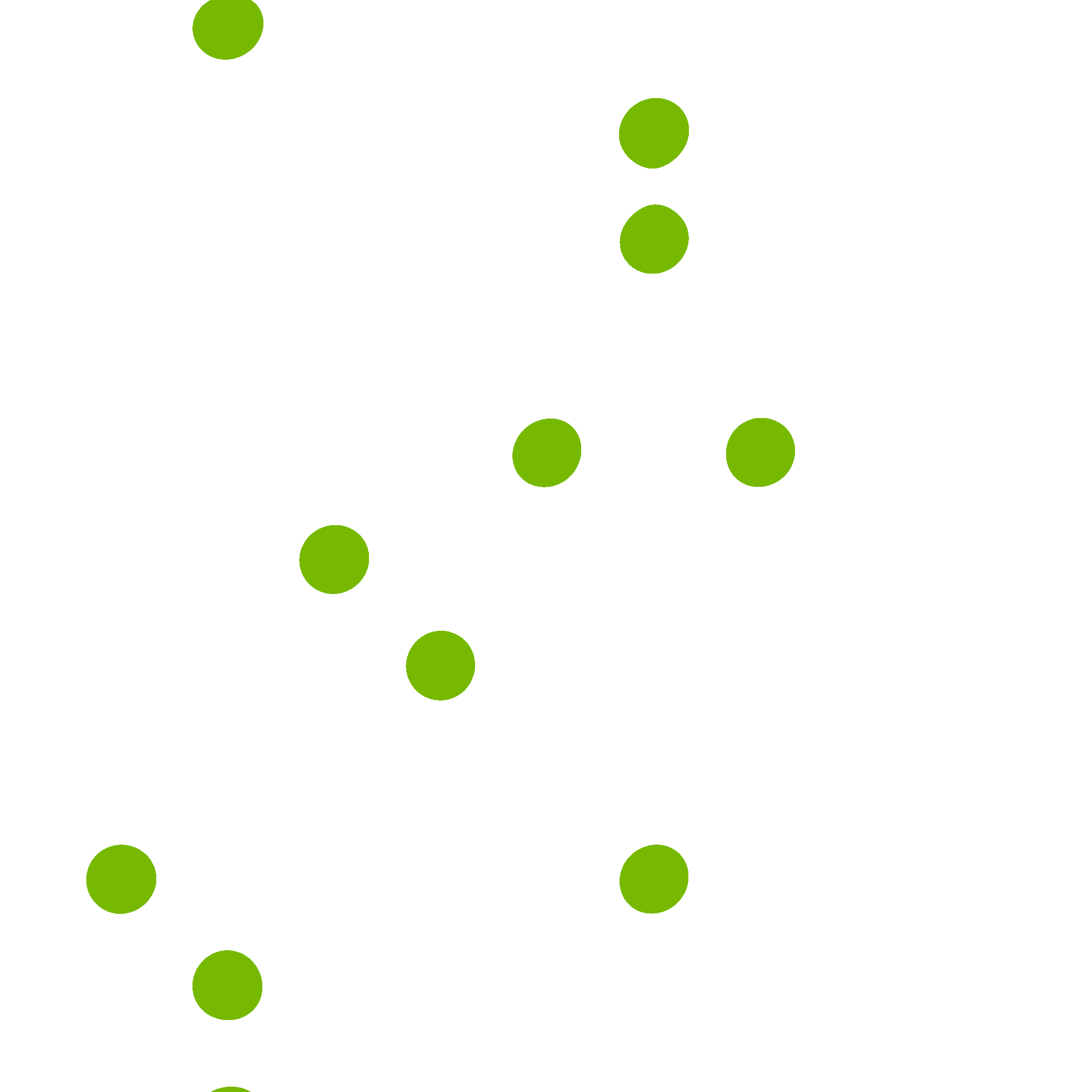} \label{fig:goodexample-iltr}}
    \caption{Good mask example for isolated shapes in the training set.}
    \label{fig:goodexample}
\end{figure}

It looks like the machine learning model knows the isolated mask shape in the training instance is bad. 
One explanation is that the model gathers the knowledge from other training instances.
Luckily, we are able to locate these training instances that contain evenly distributed isolated shapes.
As show in \Cref{fig:goodexample}, the levelset ILT engine can perform better optimization on these designs.
Thanks to the data flow in FNO, we can integrate the lithography physics in the neural network design.
This enables efficient learning of corner cases in the training set and therefore grants better mask quality.

\subsubsection{Litho-Guided Self Training:}
So far, we have shown that machine learning models, if carefully designed, are able to outperform numerical optimizer on mask optimization problems.
This motivates a training flow where the training set and the machine learning model can be updated alternatively, which is defined as \textit{litho-guided self training} (LGST).
Detailed training flow is presented in \Cref{alg:or}.
The first step is to train the machine learning model with the initial training set (line 1), where masks are generated from the ILT engine.
Following steps are $T$ rounds LGST (lines 2--12).
In each LGST round, we perform model inference on the training set and obtain the model generated masks (line 4).
Both the ML-Mask and ILT-Mask will be fed into the lithography simulation engine to measure the resist quality (lines 5--6).
Here we use MSE as a example (see definition \ref{def:mse}).
If the machine learning generated mask has better resist quality than the ILT created mask, we will replace it in the training set (lines 7--9).
At the end of $T$ rounds LGST, we will retrain the model with latest training set.

\begin{algorithm}[h]
    \caption{Litho-Guided Self Training.}
    \small
    \label{alg:or}
    \begin{algorithmic}[1]
        \Require Training dataset \{$\mathcal{Z}_{tr}$, $\mathcal{M}_{tr}$\}, LGST max iteration $T$, a random initialized machine learning mode $f(\cdot; \vec{w})$ and a lithography simulator $l(\cdot)$;
        \Ensure Trained model $f(\cdot; \vec{w})$ and updated training set \{$\mathcal{Z}_{tr}$, $\mathcal{M}_{tr}$\}.
        \State $\vec{w} \leftarrow$ Train $f$ with \{$\mathcal{Z}_{tr}$, $\mathcal{M}_{tr}$\};
        \For {$t=1,2,...,T$}
        \For {each $\vec{Z}^\ast_{tr,i} \in \mathcal{Z}_{tr}$}
        \State $\tilde{\vec{M}}_{tr,i} \leftarrow f(\vec{Z}^\ast_{tr,i}; \vec{w})$;
        \State $\textbf{MSE}_{ml} \leftarrow l(\tilde{\vec{M}}_{tr,i},\vec{Z}^\ast_{tr,i})$;
        \State $\textbf{MSE}_{ilt} \leftarrow l({\vec{M}}_{tr,i},\vec{Z}^\ast_{tr,i})$;
        \If{$\textbf{MSE}_{ml} < \textbf{MSE}_{ilt}$}
        \State $\mathcal{M}_{tr} \leftarrow$ Replace ${\vec{M}}_{tr,i}$ with $\tilde{\vec{M}}_{tr,i}$;
        \EndIf
        \EndFor
        \State $\vec{w} \leftarrow$ Train $f$ with \{$\mathcal{Z}_{tr}$, $\mathcal{M}_{tr}$\};
        \EndFor
    \end{algorithmic}
\end{algorithm}

We may also observe that LGST works in a similar manner as offline reinforcement learning \cite{offlineRL}. 
However, in the inference phase, we only need one-shot forward calculation when generating masks from design targets, 
which runs more efficiently than reinforcement learning.

\section{Experiments}
\label{sec:results}

To evaluate the effectiveness of the proposed solution, we conduct comprehensive experiments and list details in this section. 
All the experiments are conducted on a DGX platform with NVIDIA A100 GPU.

\subsection{The Dataset and Configurations}
In this paper, we adopt two groups of data set which have the same process configurations as in \cite{OPC-ICCAD2013-Banerjee}.
The legacy ILT engine used to create the training masks is the recent published LevelSet optimizer from \cite{OPC-DATE2021-Yu}, which has core computing functions implemented with CUDA.

\begin{table}[tb!]
\centering
\caption{Dataset statistics and properties.}
\label{tab:data}
\renewcommand{\arraystretch}{1.2}
\begin{tabular}{cc|ccc}
\toprule
\multicolumn{2}{c|}{Data}                              & Count & Resolution & Size  \\ \midrule
\multicolumn{1}{c|}{\multirow{2}{*}{Metal}} & Training & 1000  & 1$nm^2$/pixel        & 4$\mu m^2$  \\
\multicolumn{1}{c|}{}                       & Testing  & 10    & 1$nm^2$/pixel        & 36$\mu m^2$ \\ \midrule
\multicolumn{1}{c|}{\multirow{2}{*}{Via}}   & Training & 2784  & 1$nm^2$/pixel        & 4$\mu m^2$  \\
\multicolumn{1}{c|}{}                       & Testing  & 10    & 1$nm^2$/pixel        & 36$\mu m^2$ \\ \bottomrule
\end{tabular}
\end{table}

\begin{figure}[tb!]
	\centering 
	\includegraphics[width=.44\textwidth]{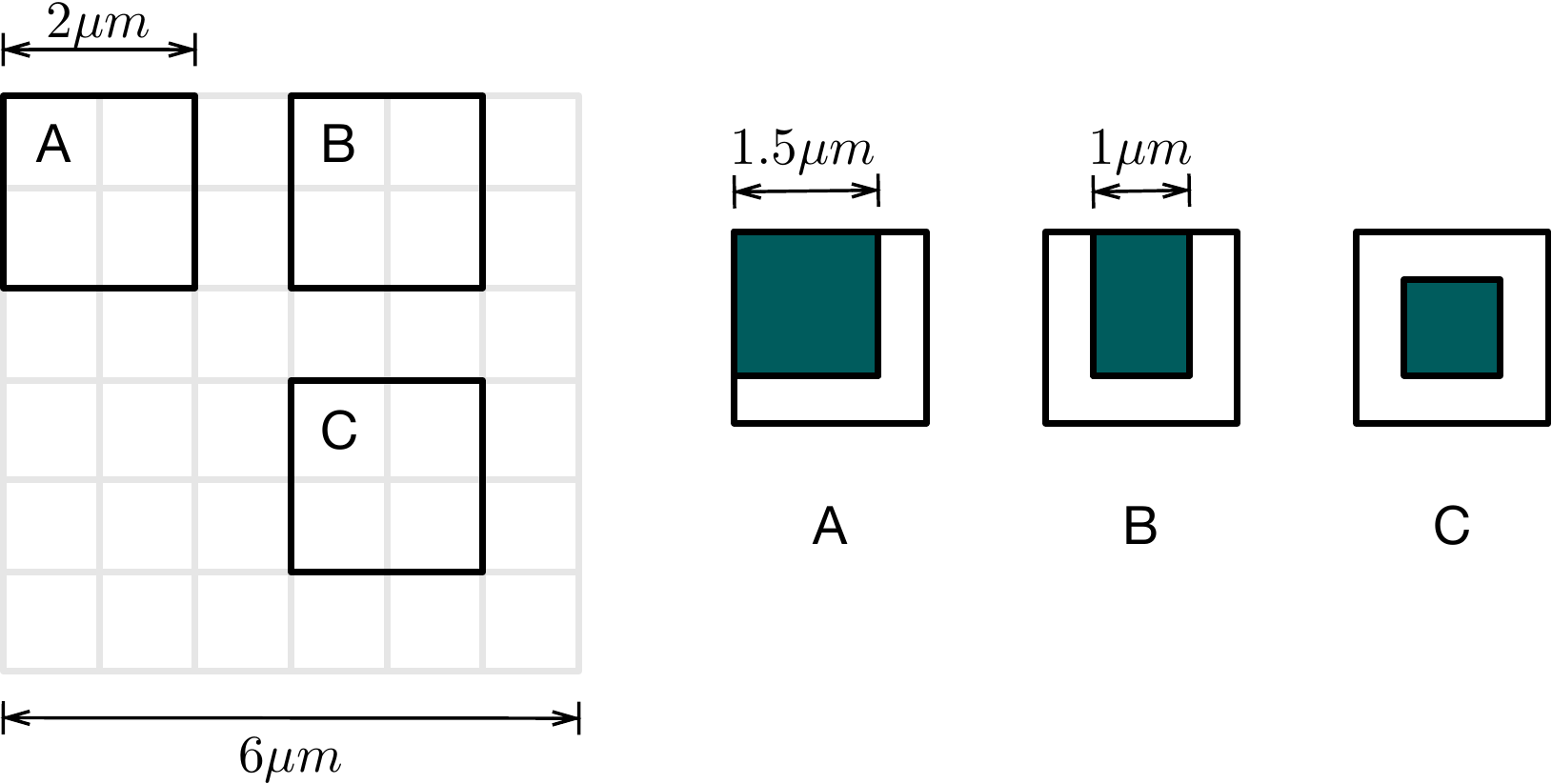}
	\caption{Large tile optimization rule. We keep different regions of the sub-tile to combine the final optimized mask.}
	\label{fig:large-opt-rule}
\end{figure}

The dataset details are listed in \Cref{tab:data}.
All designs are clipped from physical synthesised layouts and are scaled to match forward simulation engine \cite{OPC-ICCAD2013-Banerjee}.
For the metal layer designs, we have 1000 $2\mu m \times 2\mu m$ tiles used for training and ten $6\mu m \times 6\mu m$ tiles used for testing.
For the via layer designs, we have 2784 $2\mu m \times 2\mu m$ tiles used for training and ten $6\mu m \times 6\mu m$ tiles used for testing.
Training data includes target design and their corresponding masks generated from \cite{OPC-DATE2021-Yu}.
For the testing data, we employ larger tiles to demonstrate the scalability of our framework.
Our CFNO-backboned model naturally supports any-sized input.
However, the LevelSet optimizer in \cite{OPC-DATE2021-Yu} and A2-ILT \cite{OPC-DAC2022-Wang} are only applicable on $2\mu m \times 2\mu m$ tiles.
Therefore, we perform tile-based optimization on larger designs and combine the optimized tiles back to original designs.
In detail, each $6\mu m \times 6\mu m$ clip will be divided into 25 half-overlapped $2\mu m \times 2\mu m$ tiles, which will be fed into the LevelSet optimizer to create masks.
When a large tile is divided into overlapped sub-tiles, we can observe three types of sub-tiles.
We use the rule shown in \Cref{fig:large-opt-rule} to combine these sub-tiles back and preserve the boundary consistency:
\begin{itemize}
	\item Type-A: Located at the corner of the original tile. Once optimized through ILT, keep the $1.5\mu m \times 1.5\mu m$ corner region, as in the shadowed area in \Cref{fig:large-opt-rule}-A.
	\item Type-B: Located at the edge of the original tile. Once optimized through ILT, keep the $1\mu m \times 1.5\mu m$ rectangle region against the edge, as in the shadowed area in \Cref{fig:large-opt-rule}-B.
	\item Type-C: Located at the center of the original tile. Once optimized through ILT, keep the center $1\mu m \times 1\mu m$ region, as in the shadowed area in \Cref{fig:large-opt-rule}-C.
\end{itemize}
These shadowed areas will together formulate the final optimized mask of the large tile.
We also use the same rule to generate mask results from A2-ILT \cite{OPC-DAC2022-Wang}.

For each round of self training we use the same settings as in \Cref{tab:train-setting}.
Particularly, default parameters are chosen for the Adam optimizer and the loss is measured between ILT-masks and neural network generated masks.
We pick up one design for cross validation, where the MSE is used to evaluate the model performance.

\begin{table}[tb!]
\centering
\caption{Training configurations.}
\label{tab:train-setting}
\begin{tabular}{c|c}
\toprule
Configurations             & Value                \\ \midrule
Max Epoch                  & 20                   \\
Learning Rate              & 0.004                \\
Learning Rate Decay Policy & step, every 2 epochs \\
Learning Rate Decay Factor & 0.5                  \\
Batch Size                 & 16                   \\
Optimizer                  & Adam                 \\
Loss                       & L1                   \\ \bottomrule
\end{tabular}
\end{table}

\subsection{Comparison with the State-of-the-Art}
For the first experiment, we compare our framework with the state-of-the-art academic mask optimizers with details listed in \Cref{tab:result-metal} and \Cref{tab:result-via}.
Columns ``levelsetGPU \cite{OPC-DATE2021-Yu}'' corresponds to a levelset-based optimizer developed in \cite{OPC-DATE2021-Yu}.
Columns ``A2-ILT \cite{OPC-DAC2022-Wang}'' is the latest ILT engine with reinforcement learning-assisted mask condition generation.
Both ``levelsetGPU'' and ``A2-ILT'' are GPU-based optimizer. 
Columns ``MSE'' and ``PVB'' indicate the nominal resist image error and the PVB area respectively.
Columns ``EPE \#'' denotes the total number of EPE violations in the design.
Columns ``Score'' is the mask quality measurement from \cite{OPC-ICCAD2013-Banerjee} which is defined as follows:
\begin{align}
    \text{Score} &= \text{Runtime} + 5000 \times \text{EPE \#}   \nonumber \\
    &+ 4 \times \text{PVB} + 10000 \times \text{Shape Violation}.
\end{align}
Because there are no cuts and holes appearing in the results and the runtime is much smaller than other values, we only keep the ``EPE \#'' and ``PVB'' for score calculation.
For all the 20 designs, our approach achieves significantly smaller MSE and EPE violations.
This is reflected as an average EPE violation of 45.6 on metal designs compared to the 139.6 achieved by levelsetGPU \cite{OPC-DATE2021-Yu} and 128.8 achieved by A2-ILT \cite{OPC-DAC2022-Wang}.
For via designs, the advantage of our approach is even much better with 2.7 average EPE violations compared to 165.2 by levelsetGPU \cite{OPC-DATE2021-Yu} and 288.5 by A2-ILT \cite{OPC-DAC2022-Wang}.
We can also conclude that EPE and MSE are not necessarily correlated, as for some cases, A2-ILT are offering smaller MSE with much larger number of EPE violations.
PVB and MSE are usually trade-off counterparts. From the result table we can observe slightly increased PVB of our approach compared to state-of-the-arts. 
Thanks to the CFNO design and the litho-guided self training scheme, the PVB penalty is minor compared to the significant improvements of EPE violations and MSE.

\begin{table*}[tb!]
\centering
\caption{Result comparison with state-of-the-art (Metal).}
\label{tab:result-metal}
\setlength{\tabcolsep}{4pt}
\renewcommand{\arraystretch}{1.1}
\begin{tabular}{c|cccc|cccc|cccc}
\toprule
\multirow{2}{*}{Metal} & \multicolumn{4}{c|}{levelsetGPU \cite{OPC-DATE2021-Yu}}          & \multicolumn{4}{c|}{A2-ILT \cite{OPC-DAC2022-Wang}}               & \multicolumn{4}{c}{Ours}                  \\
                       & MSE      & EPE \# & PVB       & Score     & MSE      & EPE \# & PVB       & Score     & MSE      & EPE \# & PVB       & Score     \\ \midrule
1                      & 717711   & 123    & 1073631   & 4909524   & 622499   & 131    & 1206481   & 5480924   & 619400   & 42     & 1179767   & 4929068   \\
2                      & 702025   & 124    & 983446    & 4553784   & 589087   & 133    & 1101387   & 5070548   & 582073   & 33     & 1110150   & 4605600   \\
3                      & 658705   & 119    & 945891    & 4378564   & 528908   & 103    & 1073533   & 4809132   & 525663   & 28     & 1071868   & 4427472   \\
4                      & 752615   & 139    & 1077373   & 5004492   & 661972   & 169    & 1212712   & 5695848   & 667839   & 44     & 1187359   & 4969436   \\
5                      & 722932   & 151    & 1030587   & 4877348   & 607529   & 143    & 1162991   & 5366964   & 578366   & 39     & 1144289   & 4772156   \\
6                      & 614184   & 121    & 924494    & 4302976   & 492687   & 92     & 1031586   & 4586344   & 501621   & 55     & 977696    & 4185784   \\
7                      & 704913   & 142    & 1030804   & 4833216   & 591932   & 128    & 1160824   & 5283296   & 593192   & 57     & 1113514   & 4739056   \\
8                      & 783171   & 172    & 1105868   & 5283472   & 656889   & 150    & 1236872   & 5697488   & 653461   & 56     & 1234539   & 5218156   \\
9                      & 617110   & 125    & 874875    & 4124500   & 502989   & 106    & 973633    & 4424532   & 489635   & 33     & 957061    & 3993244   \\
10                     & 819572   & 180    & 1154090   & 5516360   & 642156   & 133    & 1314237   & 5921948   & 705898   & 69     & 1287713   & 5495852   \\ \midrule
Average                & 709293.8 & 139.6  & 1020105.9 & 4778423.6 & 589664.8 & 128.8  & 1147425.6 & 5233702.4 & 591714.8 & 45.6   & 1126395.6 & 4733582.4 \\
Ratio                  & 1.000        & 1.000      & \textbf{1.000}         & 1.000         & 0.831    & 0.923  & 1.125     & 1.095     & \textbf{0.834}    & \textbf{0.327}  & 1.104     & \textbf{0.991}     \\ \bottomrule
\end{tabular}
\end{table*}

\begin{table*}[tb!]
\centering
\caption{Result comparison with state-of-the-art (Via).}
\label{tab:result-via}
\renewcommand{\arraystretch}{1.1}
\begin{tabular}{c|cccc|cccc|cccc}
\toprule
\multirow{2}{*}{Via} & \multicolumn{4}{c|}{levelsetGPU \cite{OPC-DATE2021-Yu}}     & \multicolumn{4}{c|}{A2-ILT \cite{OPC-DAC2022-Wang}}              & \multicolumn{4}{c}{Ours}                 \\
                     & MSE      & EPE \# & PVB    & Score   & MSE      & EPE \# & PVB      & Score     & MSE      & EPE \# & PVB      & Score     \\ \midrule
1                    & 453635   & 124    & 278832 & 1735328 & 358447   & 140    & 335097   & 2040388   & 225608   & 3      & 318060   & 1287240   \\
2                    & 446488   & 106    & 309079 & 1766316 & 400451   & 151    & 354888   & 2174552   & 244356   & 3      & 339790   & 1374160   \\
3                    & 702076   & 182    & 404718 & 2528872 & 615320   & 276    & 495891   & 3363564   & 335072   & 2      & 467459   & 1879836   \\
4                    & 836855   & 225    & 487343 & 3074372 & 893965   & 433    & 618082   & 4637328   & 422824   & 0      & 564518   & 2258072   \\
5                    & 496560   & 130    & 329682 & 1968728 & 471114   & 212    & 390804   & 2623216   & 266208   & 9      & 368705   & 1519820   \\
6                    & 668504   & 181    & 386699 & 2451796 & 576545   & 261    & 486382   & 3250528   & 324939   & 3      & 453642   & 1829568   \\
7                    & 949451   & 232    & 637090 & 3708360 & 1114099  & 588    & 789493   & 6097972   & 563211   & 3      & 706162   & 2839648   \\
8                    & 448064   & 95     & 302426 & 1684704 & 368718   & 141    & 348087   & 2097348   & 236673   & 2      & 334928   & 1349712   \\
9                    & 609940   & 147    & 372281 & 2224124 & 534764   & 219    & 448402   & 2888608   & 298606   & 0      & 423301   & 1693204   \\
10                   & 845013   & 230    & 511550 & 3196200 & 914125   & 464    & 643113   & 4892452   & 435855   & 2      & 580563   & 2332252   \\ \midrule
Average              & 645658.6 & 165.2  & 401970 & 2433880 & 624754.8 & 288.5  & 491023.9 & 3406595.6 & 335335.2 & 2.7    & 455712.8 & 1836351.2 \\
Ratio                & 1.000        & 1.000      & \textbf{1.000}      & 1.000       & 0.968    & 1.746  & 1.222    & 1.400     & \textbf{0.519}    & \textbf{0.016}  & 1.134    & \textbf{0.754}     \\ \bottomrule
\end{tabular}
\end{table*}

We also demonstrate the efficiency of our method in \Cref{tab:time} by comparing the throughput with different mask optimizers.
Because our approach does not require further finetuning from legacy engines, we achieved the highest throughput among the three mask optimization solutions. 
In detail, we present 13890$\times$ speedup over levelsetGPU and 631$\times$ speedup over A2-ILT.
\begin{table}[tb!]
\centering
\caption{Runtime comparison.}
\label{tab:time}
\renewcommand{\arraystretch}{1.2}
\begin{tabular}{c|ccc}
\toprule
Method             & levelsetGPU \cite{OPC-DATE2021-Yu}   &A2-ILT \cite{OPC-DAC2022-Wang}&Ours             \\ \midrule
Throughput ($\mu m^2/s$)    & 0.01   &0.22 &\textbf{138.9}            \\ \bottomrule
\end{tabular}
\end{table}

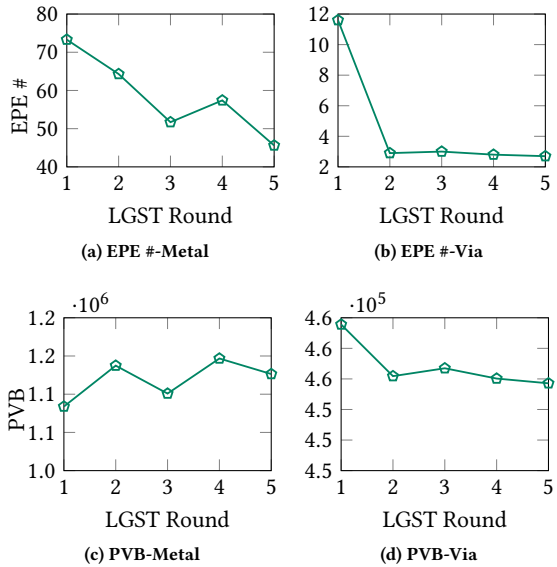
\begin{figure}
	\centering
	\subfloat[EPE \#-Metal]{\pgfplotsset{
	width =0.24\textwidth,
	height=0.2\textwidth
}
\begin{tikzpicture}[scale=1]
	\begin{axis}[minor tick num=0,
		xmin=1,
		ymin=40,
		xmax=5,
		ymax=80,
		yticklabel style={/pgf/number format/.cd, fixed, fixed zerofill, precision=0, /tikz/.cd},
		y label style={at={(axis description cs:-0.15,0.4)},rotate=0,anchor=south},
		ylabel={EPE \#},
		xlabel={LGST Round},
		xlabel near ticks,
		legend style={
			draw=none,
			at={(0.019,0.00)},
			anchor=south west,
			legend columns=1,
		}
		]
		\addplot +[line width=0.7pt] [color=NVemerald,   solid, mark=pentagon]  table [x={step},  y={epe}]  {epem.dat};
		
	\end{axis}
\end{tikzpicture}} 
	\subfloat[EPE \#-Via]{\pgfplotsset{
    width =0.24\textwidth,
    height=0.2\textwidth
}
\begin{tikzpicture}[scale=1]
\begin{axis}[minor tick num=0,
xmin=1,
ymin=2,
xmax=5,
ymax=12,
yticklabel style={/pgf/number format/.cd, fixed, fixed zerofill, precision=0, /tikz/.cd},
y label style={at={(axis description cs:-0.1,0.4)},rotate=0,anchor=south},
xlabel={LGST Round},
xlabel near ticks,
legend style={
  draw=none,
  at={(0.019,0.00)},
  anchor=south west,
  legend columns=1,
}
]
\addplot +[line width=0.7pt] [color=NVemerald,   solid, mark=pentagon]  table [x={step},  y={epe}]  {epev.dat};

\end{axis}
\end{tikzpicture}}  \\
	\subfloat[PVB-Metal]{\pgfplotsset{
	width =0.24\textwidth,
	height=0.2\textwidth
}
\begin{tikzpicture}[scale=1]
	\begin{axis}[minor tick num=0,
		xmin=1,
		ymin=1000000,
		xmax=5,
		ymax=1200000,
		yticklabel style={/pgf/number format/.cd, fixed, fixed zerofill, precision=1, /tikz/.cd},
		y label style={at={(axis description cs:-0.15,0.4)},rotate=0,anchor=south},
		ylabel={PVB},
		xlabel={LGST Round},
		xlabel near ticks,
		legend style={
			draw=none,
			at={(0.019,0.00)},
			anchor=south west,
			legend columns=1,
		}
		]
		\addplot +[line width=0.7pt] [color=NVemerald,   solid, mark=pentagon]  table [x={step},  y={epe}]  {pvbm.dat};
		
	\end{axis}
\end{tikzpicture}}   
	\subfloat[PVB-Via]{\pgfplotsset{
	width =0.24\textwidth,
	height=0.2\textwidth
}
\begin{tikzpicture}[scale=1]
	\begin{axis}[minor tick num=0,
		xmin=1,
		ymin=450000,
		xmax=5,
		ymax=460000,
		yticklabel style={/pgf/number format/.cd, fixed, fixed zerofill, precision=1, /tikz/.cd},
		y label style={at={(axis description cs:-0.1,0.4)},rotate=0,anchor=south},
		xlabel={LGST Round},
		xlabel near ticks,
		legend style={
			draw=none,
			at={(0.019,0.00)},
			anchor=south west,
			legend columns=1,
		}
		]
		\addplot +[line width=0.7pt] [color=NVemerald,   solid, mark=pentagon]  table [x={step},  y={pvb}]  {pvbv.dat};
		
	\end{axis}
\end{tikzpicture}}
	
	\caption{Litho-guided self training improves model quality.}
	\label{fig:lgst-test}
\end{figure}
\begin{figure}
	\centering
	\subfloat[Training-Metal]{\pgfplotsset{
	width =0.48\textwidth,
	height=0.3\textwidth
}
\begin{tikzpicture}[scale=1]
	\begin{axis}[minor tick num=0,
		xmin=1,
		ymin=1.5,
		xmax=20,
		ymax=7,
		yticklabel style={/pgf/number format/.cd, fixed, fixed zerofill, precision=0, /tikz/.cd},
		y label style={at={(axis description cs:-0.05,0.4)},rotate=0,anchor=south},
		ylabel={loss},
		xlabel={Epoch},
		xlabel near ticks,
		legend style={
			draw=none,
			at={(0.7,0.7)},
			anchor=south west,
			legend columns=1,
		}
		]
		\addplot +[line width=0.7pt] [color=NVemerald,   solid, mark=pentagon]  table [x={epoch},  y={loss-1}]  {trainm.dat};
		\addplot +[line width=0.7pt] [color=NVamethyst,   solid, mark=triangle]  table [x={epoch},  y={loss-5}]  {trainm.dat};
		\legend{LGST-1, LGST-5}
		
	\end{axis}
\end{tikzpicture}}  \\ 
	\subfloat[Training-Via]{\pgfplotsset{
	width =0.48\textwidth,
	height=0.3\textwidth
}
\begin{tikzpicture}[scale=1]
	\begin{axis}[minor tick num=0,
		xmin=1,
		ymin=1.5,
		xmax=20,
		ymax=7,
		yticklabel style={/pgf/number format/.cd, fixed, fixed zerofill, precision=0, /tikz/.cd},
		y label style={at={(axis description cs:-0.05,0.4)},rotate=0,anchor=south},
		ylabel={loss},
		xlabel={Epoch},
		xlabel near ticks,
		legend style={
			draw=none,
			at={(0.7,0.7)},
			anchor=south west,
			legend columns=1,
		}
		]
		\addplot +[line width=0.7pt] [color=NVemerald,   solid, mark=pentagon]  table [x={epoch},  y={loss-1}]  {trainv.dat};
		\addplot +[line width=0.7pt] [color=NVamethyst,   solid, mark=triangle]  table [x={epoch},  y={loss-5}]  {trainv.dat};
		\legend{LGST-1, LGST-5}
		
	\end{axis}
\end{tikzpicture}}
	\caption{Litho-guided self training grants faster convergence.}
	\label{fig:lgst-train}
\end{figure}
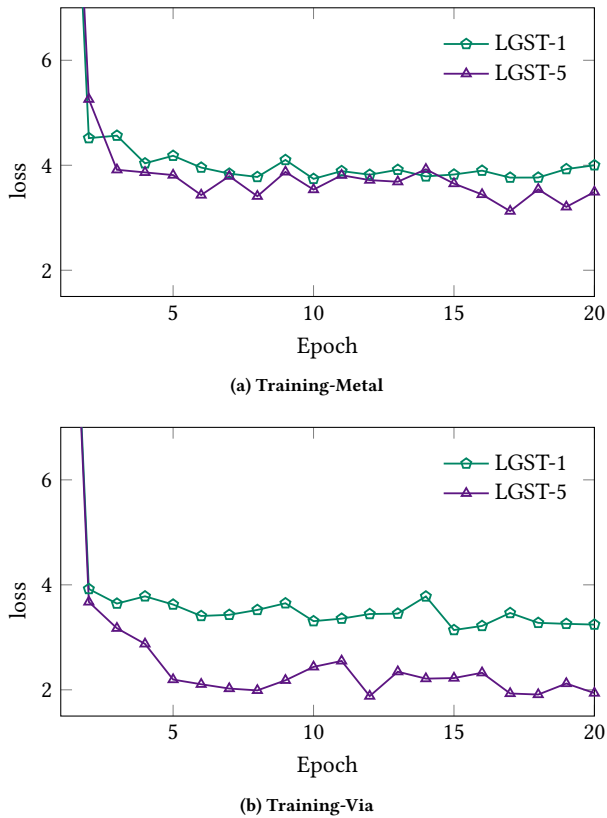

\subsection{Litho-Guided Self Training}
In the second experiment, we demonstrate the benefits of LGST.
\Cref{fig:lgst-test} shows the testing results on different LGST rounds.
Because in LGST, we update the training masks according to their quality measured in terms of MSE, 
we can observe that as LGST continues, there is a clear trend of decreasing EPE violation counts for both metal and via designs.
As for the trade-off counterpart, PVB looks stable. Although there is a slight trend of increasing PVBand area for metal designs, the penalty is still minor compared to the significant drop on EPE violations.

We also visualize the training curve on the 1st and 5th LGST rounds in \Cref{fig:lgst-train},
where we can see that as LGST continues, the model converges faster and better with lower loss.
This can be explained by the fact that we have updated a large fraction of the training set with model generated masks during LGST.
When we retrain the neural network towards these generated masks, the network is learning from itself and hence grants faster convergence.

Lastly, we statistically show how the training set are improved with LGST in \Cref{tab:lgst-dataset}, 
where each row corresponds to different LGST rounds,
columns ``Single Round'' lists the percentage of instances that are updated with better masks each round,
and columns ``Accumulated'' indicates the accumulated total number of instances that are updated compared to the original training set.
We can see from the table that in each round of LGST, a fraction of the training set will be updated with better training instances. 
However, different designs exhibit quite different LGST behaviour.
We observe that the percentage of instances that can be updated for metal layer is much smaller than via layers. 
This can be explained by the fact that metal layers are naturally more complicated and challenging than via layers.
But the trend shown in the table is still promising that the number of instances that are updated in each LGST round is keeping a 10\% updating rate.

\begin{table}[tb!]
\centering
\caption{Statistics of LGST. Each column lists the percentage of instances updated each round.}
\label{tab:lgst-dataset}
\setlength{\tabcolsep}{3pt}
\renewcommand{\arraystretch}{1.2}
\begin{tabular}{c|cc|cc}
\toprule
\multirow{2}{*}{LGST} & \multicolumn{2}{c|}{Metal} & \multicolumn{2}{c}{Via}    \\
                      & Single Round & Accumulated & Single Round & Accumulated \\ \midrule
1                     & 11.10\%      & 11.10\%     & 50.68\%      & 50.68\%     \\
2                     & 9.10\%       & 15.90\%     & 62.38\%      & 69.76\%     \\
3                     & 8.70\%       & 20.40\%     & 21.60\%      & 70.77\%     \\
4                     & 8.40\%       & 23.20\%     & 39.80\%      & 73.29\%     \\
5                     & 11.00\%      & 26.50\%     & 15.77\%      & 73.76\%     \\ \bottomrule
\end{tabular}
\end{table}

\section{Conclusion}
\label{sec:conclu}
In this paper, we focus on the problem of large scale mask optimization problem with machine learning techniques.
We propose the CFNO-backbone for efficient mask learning.
The architecture preserves the advantage of FNO for global information learning with significantly smaller computing overhead.
CFNO also supports any-sized input into our framework.
Observing several properties of MLMO problem, we propose the litho-guided self training algorithm, which gives us the opportunity to update the training set and machine learning model simultaneously.
As a result, we present the first MLMO framework that outperforms state-of-the-art academic numerical solutions in one-shot inference.

{
    \bibliographystyle{IEEEtran}
    \bibliography{ref/Top,ref/DFM,ref/Additional,ref/HSD}
}

\end{document}